\documentclass{article}

\usepackage[final]{neurips_2020}

\usepackage[utf8]{inputenc} 
\usepackage[T1]{fontenc}    
\usepackage{hyperref}       
\usepackage{url}            
\usepackage{booktabs}       
\usepackage{amsfonts}       
\usepackage{nicefrac}       
\usepackage{microtype}      

\usepackage{float}
\usepackage{graphicx}
\graphicspath{ {.} }

\usepackage{tabularx}
\usepackage[table]{xcolor}

\usepackage{wrapfig}

\title{Model Stitching: Looking For Functional Similarity Between Representations}

\author{Adriano Hernandez, Rumen Dangovski, Peter Y. Lu \& Marin Soljacic\\
MIT Physics\\
Cambridge, MA 02139, USA \\
\texttt{\{adrianoh,rumenrd,lup,soljacic\}@mit.edu}
}
\begin{document}

\maketitle

\begin{abstract}
  \textit{Model stitching} (Lenc \& Vedaldi 2015) is a compelling methodology to compare
  different neural network representations, because it allows us to measure to
  what degree they may be interchanged.
  We expand on a previous work from Bansal, Nakkiran \& Barak which used model stitching to
  compare representations of the same shapes learned by differently seeded
  and/or trained neural networks of the same architecture.
  Our contribution enables us to compare the representations learned by layers with
  different shapes from neural networks with different architectures.
  We subsequently reveal unexpected behavior of model stitching. Namely, we find that stitching, 
  based on convolutions, for small ResNets, can reach
  high accuracy if those layers come later in the first (sender) network than in
  the second (receiver), \textit{even if those layers are far apart}. 

  This leads us to hypothesize that stitches are not in fact learning to match the
  representations expected by receiver layers, but instead finding different representations which nonetheless
  yield similar results. Thus, we suggest that model stitching, naively implemented, may not necessarily always be an
  accurate measure of similarity.
\end{abstract}

\section{Introduction}
The success of deep learning for visual recognition has been attributed to the ability of neural networks to learn
good representations of their training data \cite{Rumelhart1986LearningIR}. That is, intermediate outputs (which we refer
to as ``representations'') of good neural networks 
are believed to encode meaningful information about their inputs, which these neural networks use for classification and/or other
downstream machine learning tasks \cite{goodfellow2016deep}.
However, our understanding of these representations is somewhat limited. Though
deep learning interpretability research, particularly for computer vision, has helped us
to intuitively grasp what deep neural 
networks are learning, we do not
know why good representations are learned, nor do we have a robust theory to characterize them. For example, we do not
know how to compare representations effectively.

Our goal is to improve the existing toolbox to find \textit{functional} similarity between representations.
It is not obvious how to find functional similarity, nor is it obvious exactly what it precisely
means even though we (the authors) have some intuition of it, so before we continue we provide
a sufficient definition for our purposes. For us, a representational similarity metric
is good at measuring ``functional similarity'' if we can easily use it to tell whether two
representations of one or two models are used by the model(s) for the same or similar purpose(s).
We believe this is a useful lens because if two representations can be used for similar purposes
then in some sense they encode similar information. We care about similar information
because understanding whether two models's representations
encode similar information could be useful for soft guarantees of safe, fair, or robust AI.

While we are interested in functional similarity, many papers \cite{Kornblith2019SimilarityON} 
\cite{Morcos2018InsightsOR} \cite{Ding2021GroundingRS}
look for measures of statistical or geometric similarity\footnote{
  Here, we mean simply that the representations, given proper shifts and rescales,
  are numerically or geometrically close by on average.
} because they can confirm known edge cases, like
a pair of identical representations or a representation and a vector of random noise.

These measures are a great starting point, but are not very informative in general, since different
neural networks could process and store information in ways which are analogous to eachother but not numerically
similar. In fact, they have been found \cite{Ding2021GroundingRS} \cite{Dujmovi2022SomePO} to be misleading on occasion
in both Computer Vision and Brain Sciences.

We believe that a previous work \cite{Bansal2021RevisitingMS} provides us with a potentially better functional similarity 
measure. It uses, learned transformations to \textit{translate} representations from one layer into those for another
layer. Their technique measures functional similarity, because invariant to the type of transformations used,
it tests whether two representations can be interchanged, which is a strong indicator that the two representations
function similarly. However, their work can only
compare representations with the same shapes.
We expand it to include all representations taking the form of
ResNet tensors with widths and heights that are multiples of each other.

\section{Experimental Setup}
\label{ExperimentalSetup}
\subsection{Models and Dataset}
We compare all different
layers of all ResNets with a number of layers ranging from ten to eighteen.
These ResNets are trained on CIFAR-10 for comparable results
with Bansal et. al. These small ResNets we characterize with 4-tuples, where 
each element is either one or two, representing the number of residual blocks per stage\footnote{
   Residual blocks are partitioned into four stages of consecutive blocks, within which they have the same shape. At each stage, the
   width and height halve, while the width doubles \cite{He2016DeepRL}.
}. Since we use at most two blocks per stage, we can denote these 4-tuples unambiguously as
$R_{1111}$, $R_{1112}$, and so on. There are $2^4 = 16$ such ResNets. Note that $R_{2222}$ is equivalent
to the well-known Resnet18 architecture, while $R_{1111}$ is equivalent to Resnet10.

\subsection{Experiment}
We train each possible Small Resnet on CIFAR-10, yielding an accuracy above 90\%. We also
generate a randomly initialized, untrained network for each Small Resnet architecture and confirm
that these have an accuracy of around 10\%\footnote{There are ten classes in CIFAR-10.}. All these
networks are frozen and cannot learn during stitching.

We stitch every ordered pair of Small Resnets. There are $16 \cdot 16 = 256$ such pairs.
In every ordered pair of networks being stitched, the former is called the \textit{sender}, and the latter
is the \textit{receiver}. A stitch is used to transform the output of the sender at an intermediate layer
before inputing it into an intermediate layer in the reciever. For any network $A$ we can consider 
layer $i$ as $A_i$, the first $i$  layers (assuming we start at zero) as $A_{<i}$, and the layers 
after $i$ as $A_{i<}$. If we wish to include layer $i$ we can always call such (partial) networks
$A_{\leq i}$ or $A_{i\leq}$. For an input $x$, if we call the sender $A$, the reciever $B$,
and the stitch $S$, we call $C = B_{j<}(S(A_{\leq i}(x)))$ the \textit{stitched network}. Normally,
we train $S$ by doing backpropagation on the stitched network with both $A$ and $B$ frozen. The
resulting accuracy is used to find the similarity\footnote{
  We use the downstream accuracy of the stitched network as our similarity measure for simplicity,
  since the original networks' accuracies were over 90\%. Otherwise, a ratio or difference, as used
  in Bansal et. al. may be more informative. In our case, no choice amongst these would change
  our qualitative result.
} between $A_{\leq i}(x)$ and $B_{\leq j}(x)$, the
former of which is called the \textit{provided representation} and the latter of which is called
the \textit{expected representation}.

Unlike Bansal et al., which only compare corresponding blocks (i.e. $i = j$) of a sender and
reciever with the same architecture,
we stitch from \textit{all} residual blocks of the sender into
\textit{all} residual blocks of the receiver even when they have different architectures, as long as
they are Small Resnets. Also unlike Bansal et al., we only vary
our neural networks by their initialization weights, but our setup is otherwise nearly identical. To
be able to stitch between all blocks, we use strided convolutions or upsampling when the dimensions
are not the same.
In our case the heights and widths vary by powers of two so we can simple use 2x2, 4x4,
or similarly-sized convolutions to downsample. For upsampling we use 2x, 4x, or similarly
sized 2D nearest-upsampling (meaning that an element is copied into a grid of equally-valued elements)
before applying a 1x1 convolution in the stitch.

We use the randomly-initialized ResNets as controls. Our controls enable us to make sure that
the stitches are appropriately powerful. By powerful we mean how complex the functions are that
stitches can represent. If a stitch is very powerful, then even with random networks it should be able to
yield high downstream accuracy because it can learn any transformation. In this case our 
stitches would always yield high ``similarity'' and therefore be uninformative.
We can be sure this is not the case by ensuring that
the stitches never yield high similarity for random networks (where only overly powerful stitches
would).

\begin{center}
  \begin{figure}[H]
     \centering
     \includegraphics[width=12cm]{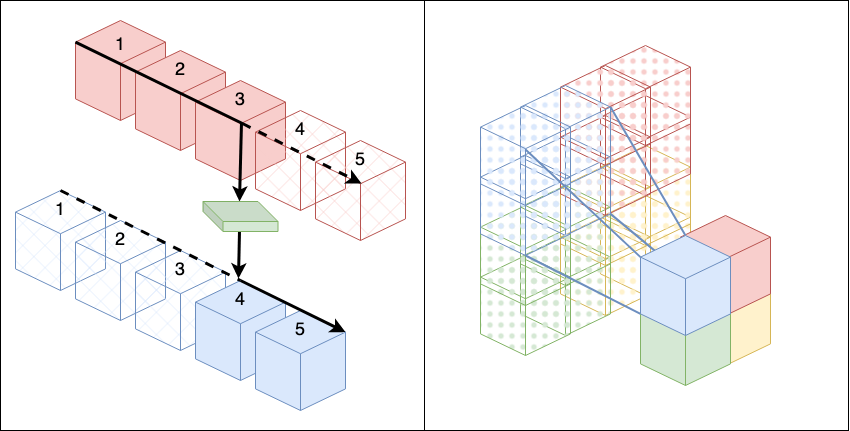}
     \caption{\textbf{On the left}, a diagram exemplifying a stitch from the red (sender) network into the blue (reciever) 
     network comparing layer 3 from both. In this diagram, the blue layer 3 is the expected representation. Unused layers
     in the stitched network are displayed as partially translucent. The stitch
     is depicted in green. The arrows denote the flow of computation. The dashed arrows denote the flow of computation
     in regular operation, absent of stitching. \textbf{On the right}, a diagram exemplifying a 2x2 convolution (downsampling)
     from left (dotted) to right (solid). A 2x2 convolution such as this one could be used to stitch from a representation with 
     larger width and height to one of smaller dimensions. The colors are  chosen so as to elucidate which elements correspond.
     The blue lines further highlight the correspondence for the blue elements.
     In the case of upsampling, the image can be read from right (solid) to left (dotted), where the solid blue 
     element is copied four times to the
     dotted blue elements before it is used (later) for a 1x1 convolution.}
  \end{figure}
\end{center}

\section{Results}
\label{Results}
For every ordered pair of networks, we plot the accuracy of all the stitched networks on a grid,
based on the sender's layer and the receiver's layer. The layer is
denoted by an integer which counts how many residual blocks came before it\footnote{
   The initial convolution is ``0,'' the first block of the first stage is ``1,'' and so on.
}. The value in the grid element is the accuracy of the stitched network after traning. Thus, this
grid is a \textit{similarity matrix} where entry $i, j$ corresponds to the similarity of
$A_{\leq i}(x)$ and $B_{\leq j}$.

\subsection{Expectations}
We hypothesized that for similarity matrices between networks of the same architecture we would
see a high similarity diagonal. For networks of different architectures we hoped for a shorter diagonal
(since the matrix isn't square) or a diagonal with a different slope.

Generally, however, we assumed
that there would exist some non-negligible number $\epsilon$ such that if we mapped each
sender layer's representation to its most similar counterpart in the reciever,
and their similarity was $s$, the similarity between that layer's representation 
in the sender and every other layer's in the reciever would be
less than $s - \epsilon$. Moreover, we assumed that such a mapping would be injective. Intuitively,
we thought that there would be a one-to-one correspondence between most layers in the sender
to most layers in the reciever. We did not expect any layers' representation in the sender to have 
high similarity to \textit{multiple} layers' representations in the reciever.

These hypotheses make sense because Bansal et. al.'s findings suggest that each layer in the sender
should have at least one similar layer in the reciever---at least in the cast of identical architectures,
where those two layers are the corresponding ones; and it is usually assumed that distant layers
have different information, and so they should not be similar. However, \textbf{we found that every
layer in the sender was extremely similar to all layers before it in the reciever by proportion of
neural network length}. That is to say, for small Resnets,
regardless of the architecture, if we had one Resnet of length $I$ and another of length $J$, then if
$\frac{j}{J} \leq \frac{i}{I}$ the similarity was high between layer $i$ in the $I$-length network and 
layer and $j$ in the $J$-length network. Visually, this looks
like a triangle in the lower left-hand corner of the similarity matrix. The triangle's endoints are the top left cell,
the bottom left cell, and the bottom right cell. This pattern is visible in the figure below and quite perplexing.

For our controls we expected to see low stitching accuracy throughout the board since the networks are random,
and we did. With that said, some of the top left or bottom right elements have high similarity depending on
whether the sender or reciever was random. In the case of a random sender and trained reciever, if $i$ and $j$
are small, it is easy for $S$ to undo $A_{\leq i}$, and give $B_{j<}$ something usable. The opposite
case is analogous.

\begin{center}
   \begin{figure}[H]
      \centering
      \includegraphics{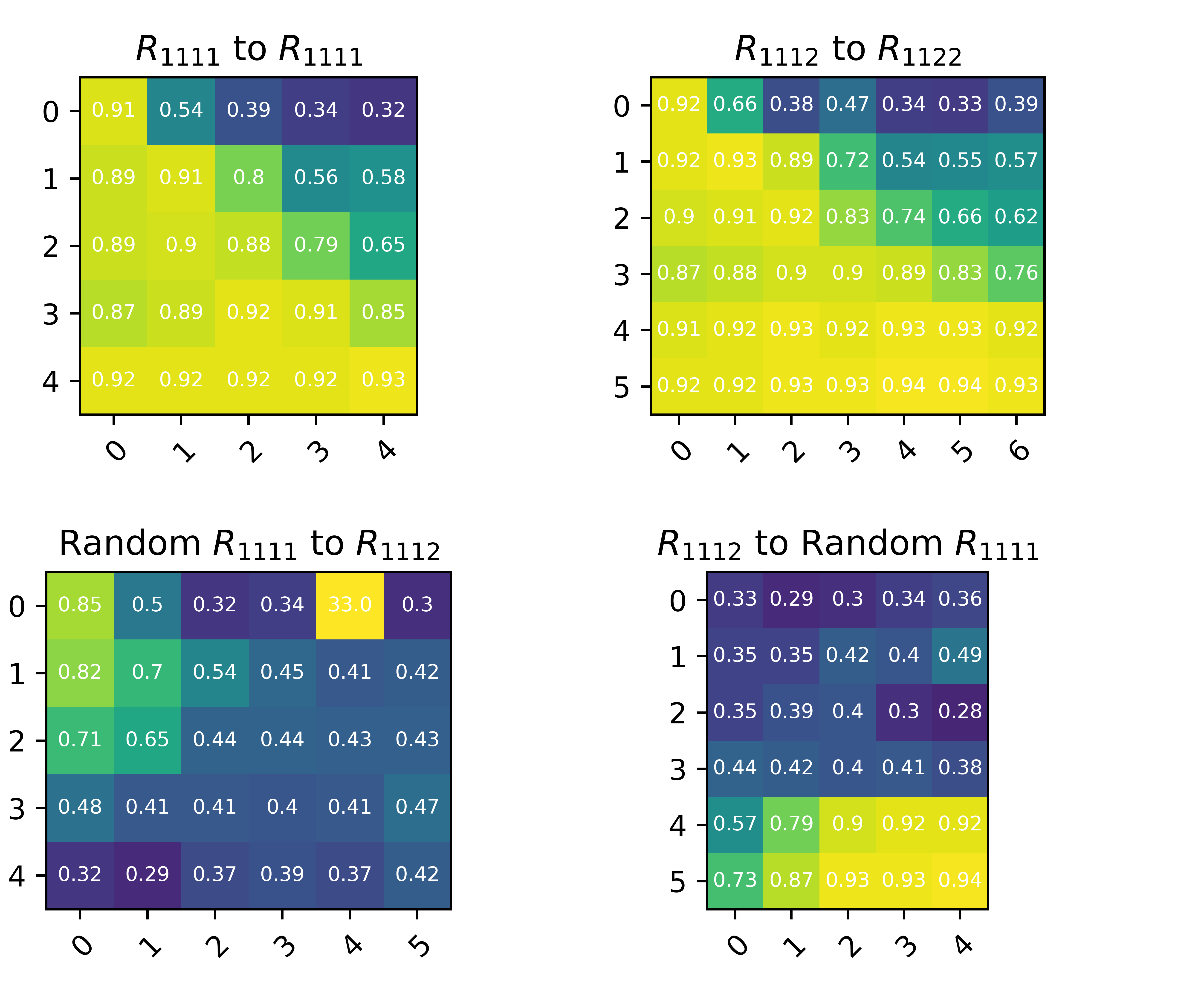}
      \caption{Triangle similarity pattern between trained Small Resnets and (expected) low similarity pattern for
      random ones. The plot is to be interpreted as a similarity matrix from sender to reciever. This is indicative
      of the pattern we saw on all such Resnets.}
   \end{figure}
\end{center}

\subsection{Conclusion}
The most interesting aspect of our results is the high accuracy of the stitching network
for layers in the lower left hand triangle. It is true that our findings do \textit{not} contradict Bansal et. al.'s
findings since they \textit{only stitched on the diagonal} which yielded the same high accuracy for them
as it did for us. However, we are still surprised. Given that we interpret the stitching network accuracy
as a similarity, our results suggest that each sender representation is similar with \textit{all}
the receiver representations from a layer (proportionally) before it. We expected to see that each layer would be
similar to a couple (nearby) layers at most because the standard narrative has been that 
every layer loses some amount of granular information, and so that information should not be reconstructable
in an \textit{interchangeability} test like stitching.

In our conclusions we focus our scope on the fact that layers are similar to those before them for similar, but not 
equal, length networks, 
instead of the fact that those layers ``before them'' need only be proportionally (to their own network's length) before, 
since most of our networks did not vary in length very much.
However, the poportionality finding could prove to be interesting to explore in future work, since it may tell us something
about Resnet length-invariant properties. A slightly more extreme case for the curious in depicted in the Appendix.

We see two main explanations for our results. The first is that the common narrative could be wrong and some neural
networks may in fact be able to maintain most if not all of the granular information of the image throughout their
processing of it. An alternative, but not uncommon, narrive to the standard one is that neural networks progressively 
discard information from layer to layer until only the class information remains. In that view, our results are actually
expected since the stitch need only make sure that it transfers the class information. This could make sense since nearest
upsampling is lossless, while downsampling is lossy, which matches the observation that stitching backwards yields high
accuracy but not forwards, suggesting that it is easier to retain the information in the backwards direction.

The second explanation we see is that the stitch may be able to give the reciever a representation which,
despite being different from that which is expected, nonetheless yields high accuracy. Intuitively, the stitch may be
able to figure out how to generate some generic, albeit unrealistic, set of the most salient features for the
recieving layer to classify in a given way. Perhaps it is generating the ``average'' human, or some sort of adversarial
representation. 

To rigorously determine why our obervations are as they are, however, a deeper analysis is required.
In the Appendix, we further discuss sanity tests we executed to try and buttress our second explanation, which
we henceforth dub the \textit{hacking} hypothesis.

Despite our difficulty generalizing model stitching, we still see it as an important
step forward in our ability to compare representations, since its focus on
\textit{functional} similarity makes its results more salient than those from
geometric closeness or statistical measures. Unlike existing measures of similarity which
tend to look for literal distance between representations, stitching follows a process whereby
we define what behaviors should be exhibited by similar representations (i.e. they should
be interchangeable up to a degree of flexibility determined by the function class of the stitch)
and then devise tasks/experiments that test these behaviors on the representations under question.
This approach is far better, because the research community understands tasks much better than
the numerical, geometric, or statistical properties of representations. Moreover, it is
easier to interpret the resulting similarity measurements in terms of accuracy or other such
\textit{functional} quantities, making techniques like stitching more useful, even in practice.
We hope to see more
representational comparison techniques following the high-level process outlined above in the future.

\begin{ack}
This work was partially funded by MIT SuperUROP as well as by the National Science Foundation under Cooperative Agreement PHY-2019786 (The NSF AI Institute for Artificial Intelligence and Fundamental Interactions, http://iaifi.org/).
\end{ack}


\newpage
\appendix
\section{Appendix}
Here we provide additional information that may be helpful to readers, but which was
beyond the scope of the main paper. There are four subsections: further details on our
testing procedures; results from larger Resnets to support the idea that our surprising findings
could generalize; numerical sanity testing to try and confirm or disprove the \textit{hacking}
hypothesis; and image-generation to try and visualize the possibility of the \textit{hacking}
hypothesis.
\subsection*{Experimental Details}
We train our stitches for four epochs with momentum 0.9,
batch size 256, weight decay 0.01, learning rate 0.01, and 
a post-warmup cosine learning rate scheduler. We chose our
hyperparameters because they were effective for training the
Small Resnets between which we stitched. Below is an example of
a Small Resnet for clarity on our architecture.
\begin{center}
  \begin{figure}[H]
     \centering
     \includegraphics[width=3cm]{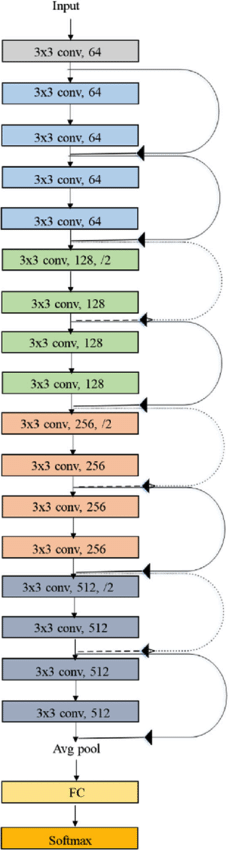}
     \caption{Resnet18 \cite{He2016DeepRL}, equivalent to $R_{2222}$
     using our nomenclature.}
  \end{figure}
\end{center}
\subsection*{Extrapolation to Larger Resnets}
\begin{center}
  \begin{figure}[H]
     \centering
     \includegraphics[width=14cm]{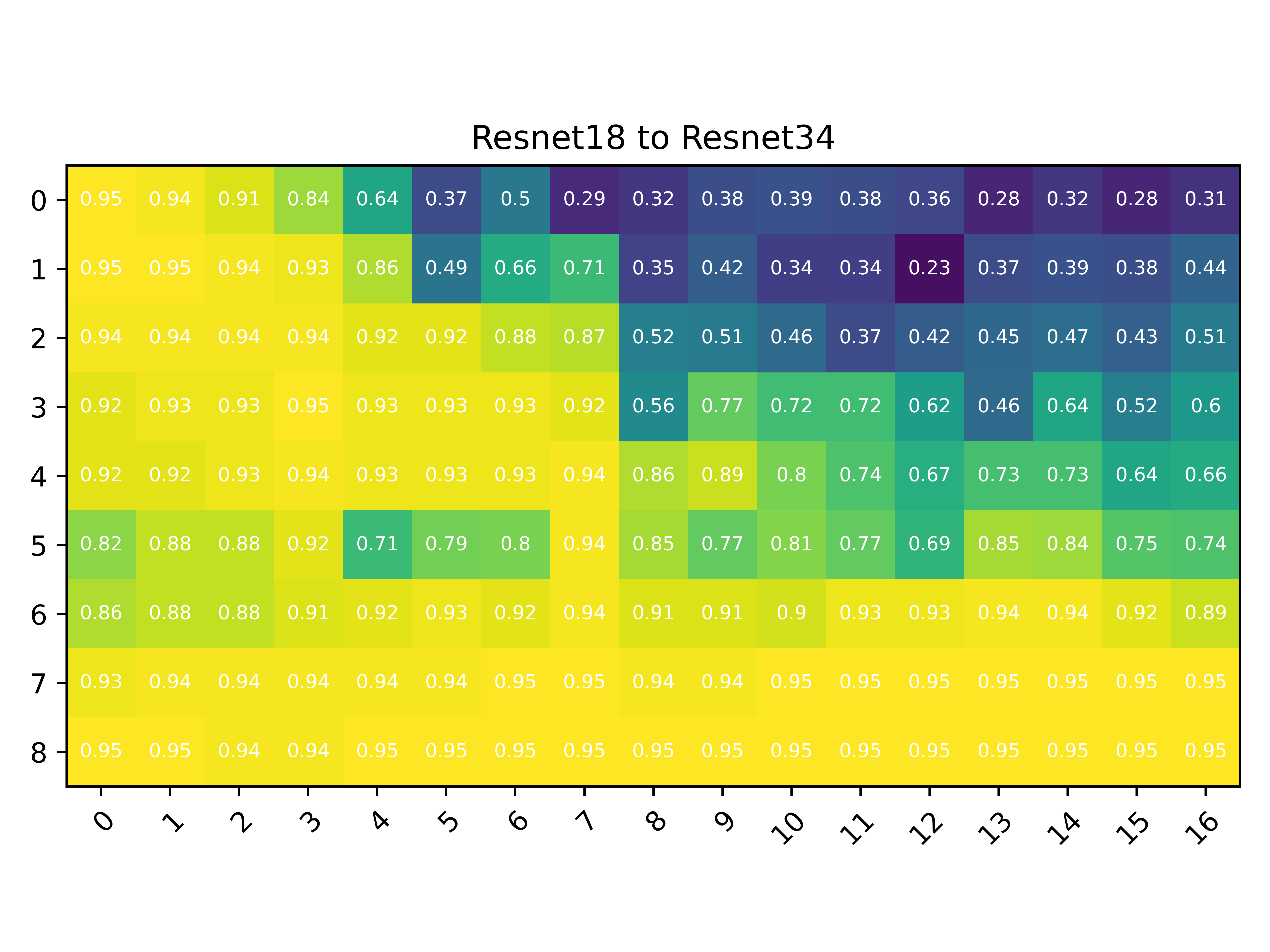}
     \caption{Our results generalize to larger Resnets of a similar type (also on CIFAR-10).
     We were able to yield the same triangular pattern between Resnet18 and Resnet34 in both
     directions, suggesting that our results are the function of some general property of the
     Resnet architecture or of CIFAR-10. One thing that becomes more aparent for Resnet
     pairs whose lengths are very different, is that it is not whether the reciever layer comes
     before the sender layer, numerically, that matters, but whether it comes before it proportional
     to the length of the reciever Resnet. This is perplexing. Visually, in the table this means
     that the lower left hand triangle of high similarity includes all elements below the diagonal,
     from the top left cell to the bottom right cell, \textit{regardless of the slope of this diagonal}.}
  \end{figure}
\end{center}
\subsection*{Numerical Sanity Testing}
When we found that our metric was finding high similarity between distant layers, we decided to sanity test
this result using numerical testing. We averaged the mean squared error between three pairs of values:
the expected representation with that generated by a vanilla stitch (trained with backpropagation on the CIFAR-10
classification task, as discussed in the body of this paper); the expected representation with that generated 
by a similarity-trained stitch; and the representation generated by a similarity-trained stitch with that generated 
by a vanilla stitch. Respectively, we refer to these three pairs as \textbf{EV}, \textbf{ES}, and \textbf{SV}.
The mean squared error is over the elements in the representations' tensors. For those mean squared errors, we
found the minimum, mean, maximum, and standard deviation over the (cartesian product of the) entire dataset of
CIFAR-10 and all the pairs of layers across \textit{all} Small Resnets. We also measured the same statistics
for only corresponding layers (i.e. layer one with layer one) to get a baseline similar to what Bansal
et. al.'s work may have yielded.

Unlike the vanilla stitch, the similarity-trained stitch was trained to minimize the mean squared error between
the expected representation of the reciever (at a layer) and the output of the corresponding stitch. For example,
consider the input $x$, sender $A$, reciever $B$, and stitch $S$. Consider the stitched
network $C = B_{j<}(S(A_{\leq i}))$, and recall that the expected representation is $B_{\leq j}$ (that is, the computation
up to, including, layer $j$ of the reciever). The vanilla stitch would be trained using backpropagation on $C$ with
all the weights of $A$ and $B$ frozen (only the weights in $S$ can change). However, the similarity-trained stitch
would be trained on $(S(A_{<i}) - B_{\leq j})^2$. The purpose of the similarity-trained stitch is to get a baseline for
what a ``low'' mean squared error is. We include its task accuracy at the end of this section as a curiosity.

Below we plot our results in two tables. We denote the highest difference with red, that with the second highest 
difference with yellow, and that with the lowest difference with green. We expect, therefore, to see red in the
\textbf{EV} and \textbf{SV} columns and green in the \textbf{ES} column. That is because the similarity-trained
stitch should be closer to the expected representation (since the their difference is its loss function) than
the vanilla stitch is to either. If the vanilla stitch is learning to hack the reciever then we expect its
difference to be larger by many orders of magnitude than the similarity-trained stitch. While we do see that
our prediction is typically correct, the difference is not as large nor as decimatingly common as we had hoped,
and so we cannot conclude, from these results, that the stitch is likely hacking the reciever.
\begin{table}[ht!]
   \small
   \caption{Diagonals\strut}
   \scriptsize
   \begin{tabularx}\linewidth{||X|X|X||X|X|X||X|X|X||X|X|X|| }
      \hline
      \multicolumn{3}{||c||}{Minimum} & \multicolumn{3}{c||}{Mean} & \multicolumn{3}{c||}{Maximum} & \multicolumn{3}{c||}{Standard Deviation} \\
      \hline
      \textbf{EV}&\textbf{ES}&\textbf{SV}&\textbf{EV}&\textbf{ES}&\textbf{SV}&\textbf{EV}&\textbf{ES}&\textbf{SV}&\textbf{EV}&\textbf{ES}&\textbf{SV} \\
      \hline
      \cellcolor[HTML]{F8CECC}{\color[HTML]{000000}2.0e-3}&\cellcolor[HTML]{D5E8D4}{\color[HTML]{000000}2.2e-5}&\cellcolor[HTML]{FFF2CC}{\color[HTML]{000000}1.2e-3} & \cellcolor[HTML]{F8CECC}{\color[HTML]{000000}4.4e-2}&\cellcolor[HTML]{D5E8D4}{\color[HTML]{000000}1.5e-2}&\cellcolor[HTML]{FFF2CC}{\color[HTML]{000000}2.3e-2} & \cellcolor[HTML]{F8CECC}{\color[HTML]{000000}2.8e-1}&\cellcolor[HTML]{FFF2CC}{\color[HTML]{000000}1.9e-1}&\cellcolor[HTML]{D5E8D4}{\color[HTML]{000000}1.1e-1} & \cellcolor[HTML]{F8CECC}{\color[HTML]{000000}6.1e-2}&\cellcolor[HTML]{D5E8D4}{\color[HTML]{000000}3.7e-2}&\cellcolor[HTML]{FFF2CC}{\color[HTML]{000000}3.9e-2} \\
      \hline
      \cellcolor[HTML]{F8CECC}{\color[HTML]{000000}1.3e-1}&\cellcolor[HTML]{FFF2CC}{\color[HTML]{000000}5.5e-2}&\cellcolor[HTML]{D5E8D4}{\color[HTML]{000000}2.7e-3} & \cellcolor[HTML]{F8CECC}{\color[HTML]{000000}4.3e+5}&\cellcolor[HTML]{D5E8D4}{\color[HTML]{000000}7.6e+4}&\cellcolor[HTML]{FFF2CC}{\color[HTML]{000000}1.7e+5} & \cellcolor[HTML]{F8CECC}{\color[HTML]{000000}4.2e+6}&\cellcolor[HTML]{D5E8D4}{\color[HTML]{000000}3.6e+5}&\cellcolor[HTML]{FFF2CC}{\color[HTML]{000000}3.7e+6} & \cellcolor[HTML]{F8CECC}{\color[HTML]{000000}1.1e+6}&\cellcolor[HTML]{D5E8D4}{\color[HTML]{000000}1.3e+5}&\cellcolor[HTML]{FFF2CC}{\color[HTML]{000000}7.3e+5} \\
      \hline
      \cellcolor[HTML]{F8CECC}{\color[HTML]{000000}1.5e-2}&\cellcolor[HTML]{D5E8D4}{\color[HTML]{000000}3.5e-5}&\cellcolor[HTML]{FFF2CC}{\color[HTML]{000000}7.9e-3} & \cellcolor[HTML]{F8CECC}{\color[HTML]{000000}3.5e-1}&\cellcolor[HTML]{D5E8D4}{\color[HTML]{000000}1.1e-3}&\cellcolor[HTML]{FFF2CC}{\color[HTML]{000000}3.2e-1} & \cellcolor[HTML]{FFF2CC}{\color[HTML]{000000}5.3e-1}&\cellcolor[HTML]{D5E8D4}{\color[HTML]{000000}5.9e-3}&\cellcolor[HTML]{F8CECC}{\color[HTML]{000000}5.3e-1} & \cellcolor[HTML]{FFF2CC}{\color[HTML]{000000}1.6e-1}&\cellcolor[HTML]{D5E8D4}{\color[HTML]{000000}2.0e-3}&\cellcolor[HTML]{F8CECC}{\color[HTML]{000000}1.6e-1} \\
      \hline
      \cellcolor[HTML]{F8CECC}{\color[HTML]{000000}1.6e-1}&\cellcolor[HTML]{D5E8D4}{\color[HTML]{000000}1.7e-2}&\cellcolor[HTML]{FFF2CC}{\color[HTML]{000000}1.2e-1} & \cellcolor[HTML]{F8CECC}{\color[HTML]{000000}1.3e+2}&\cellcolor[HTML]{D5E8D4}{\color[HTML]{000000}4.9e+0}&\cellcolor[HTML]{FFF2CC}{\color[HTML]{000000}1.2e+2} & \cellcolor[HTML]{FFF2CC}{\color[HTML]{000000}1.4e3}&\cellcolor[HTML]{D5E8D4}{\color[HTML]{000000}6.0e+1}&\cellcolor[HTML]{F8CECC}{\color[HTML]{000000}1.4e+3} & \cellcolor[HTML]{FFF2CC}{\color[HTML]{000000}2.9e+2}&\cellcolor[HTML]{D5E8D4}{\color[HTML]{000000}1.3e+1}&\cellcolor[HTML]{F8CECC}{\color[HTML]{000000}2.9e+2}  \\
      \hline
   \end{tabularx}
   \medskip
   \small
   \caption{All Stitches\strut}
   \scriptsize
   \begin{tabularx}\linewidth{||X|X|X||X|X|X||X|X|X||X|X|X|| }
      \hline
      \multicolumn{3}{||c||}{Minimum} & \multicolumn{3}{c||}{Mean} & \multicolumn{3}{c||}{Maximum} & \multicolumn{3}{c||}{Standard Deviation} \\
      \hline
      \textbf{EV}&\textbf{ES}&\textbf{SV}&\textbf{EV}&\textbf{ES}&\textbf{SV}&\textbf{EV}&\textbf{ES}&\textbf{SV}&\textbf{EV}&\textbf{ES}&\textbf{SV} \\
      \hline
      \cellcolor[HTML]{F8CECC}{\color[HTML]{000000}1.6e-3}&\cellcolor[HTML]{D5E8D4}{\color[HTML]{000000}5.4e-6}&\cellcolor[HTML]{FFF2CC}{\color[HTML]{000000}7.8e-4} & \cellcolor[HTML]{F8CECC}{\color[HTML]{000000}4.5e-2}&\cellcolor[HTML]{D5E8D4}{\color[HTML]{000000}1.6e-2}&\cellcolor[HTML]{FFF2CC}{\color[HTML]{000000}2.0e-2} & \cellcolor[HTML]{F8CECC}{\color[HTML]{000000}3.6e+0}&\cellcolor[HTML]{FFF2CC}{\color[HTML]{000000}1.3e+0}&\cellcolor[HTML]{D5E8D4}{\color[HTML]{000000}5.9e-1} & \cellcolor[HTML]{F8CECC}{\color[HTML]{000000}1.4e-1}&\cellcolor[HTML]{FFF2CC}{\color[HTML]{000000}5.9e-2}&\cellcolor[HTML]{D5E8D4}{\color[HTML]{000000}3.3e-2} \\
      \hline
      \cellcolor[HTML]{F8CECC}{\color[HTML]{000000}1.6e-4}&\cellcolor[HTML]{D5E8D4}{\color[HTML]{000000}1.7e-7}&\cellcolor[HTML]{FFF2CC}{\color[HTML]{000000}1.5e-4} & \cellcolor[HTML]{F8CECC}{\color[HTML]{000000}2.6e+5}&\cellcolor[HTML]{FFF2CC}{\color[HTML]{000000}1.1e+5}&\cellcolor[HTML]{D5E8D4}{\color[HTML]{000000}6.3e+4} & \cellcolor[HTML]{F8CECC}{\color[HTML]{000000}1.8e+7}&\cellcolor[HTML]{FFF2CC}{\color[HTML]{000000}9.2e+6}&\cellcolor[HTML]{D5E8D4}{\color[HTML]{000000}5.8e+6} & \cellcolor[HTML]{F8CECC}{\color[HTML]{000000}1.0e+6}&\cellcolor[HTML]{FFF2CC}{\color[HTML]{000000}4.3e+5}&\cellcolor[HTML]{D5E8D4}{\color[HTML]{000000}3.4e+5} \\
      \hline
      \cellcolor[HTML]{F8CECC}{\color[HTML]{000000}1.4e-2}&\cellcolor[HTML]{D5E8D4}{\color[HTML]{000000}6.3e-6}&\cellcolor[HTML]{FFF2CC}{\color[HTML]{000000}7.9e-3} & \cellcolor[HTML]{FFF2CC}{\color[HTML]{000000}2.2e+0}&\cellcolor[HTML]{D5E8D4}{\color[HTML]{000000}2.1e-3}&\cellcolor[HTML]{F8CECC}{\color[HTML]{000000}2.2e+0} & \cellcolor[HTML]{FFF2CC}{\color[HTML]{000000}2.3e+2}&\cellcolor[HTML]{D5E8D4}{\color[HTML]{000000}1.9e-2}&\cellcolor[HTML]{F8CECC}{\color[HTML]{000000}2.3e+2} & \cellcolor[HTML]{FFF2CC}{\color[HTML]{000000}1.2e+1}&\cellcolor[HTML]{D5E8D4}{\color[HTML]{000000}4.3e-3}&\cellcolor[HTML]{F8CECC}{\color[HTML]{000000}1.2e+1} \\
      \hline
      \cellcolor[HTML]{F8CECC}{\color[HTML]{000000}1.3e-1}&\cellcolor[HTML]{D5E8D4}{\color[HTML]{000000}1.6e-2}&\cellcolor[HTML]{FFF2CC}{\color[HTML]{000000}9.4e-2} & \cellcolor[HTML]{F8CECC}{\color[HTML]{000000}7.3e+4}&\cellcolor[HTML]{FFF2CC}{\color[HTML]{000000}4.2e+4}&\cellcolor[HTML]{D5E8D4}{\color[HTML]{000000}2.1e+4} & \cellcolor[HTML]{F8CECC}{\color[HTML]{000000}1.2e+7}&\cellcolor[HTML]{FFF2CC}{\color[HTML]{000000}6.9e+6}&\cellcolor[HTML]{D5E8D4}{\color[HTML]{000000}4.8e+6} & \cellcolor[HTML]{F8CECC}{\color[HTML]{000000}5.9e+5}&\cellcolor[HTML]{FFF2CC}{\color[HTML]{000000}3.4e+5}&\cellcolor[HTML]{D5E8D4}{\color[HTML]{000000}2.1e+5} \\
      \hline
   \end{tabularx}
\end{table}
\begin{center}
  \begin{figure}[H]
     \centering
     \includegraphics[width=14cm]{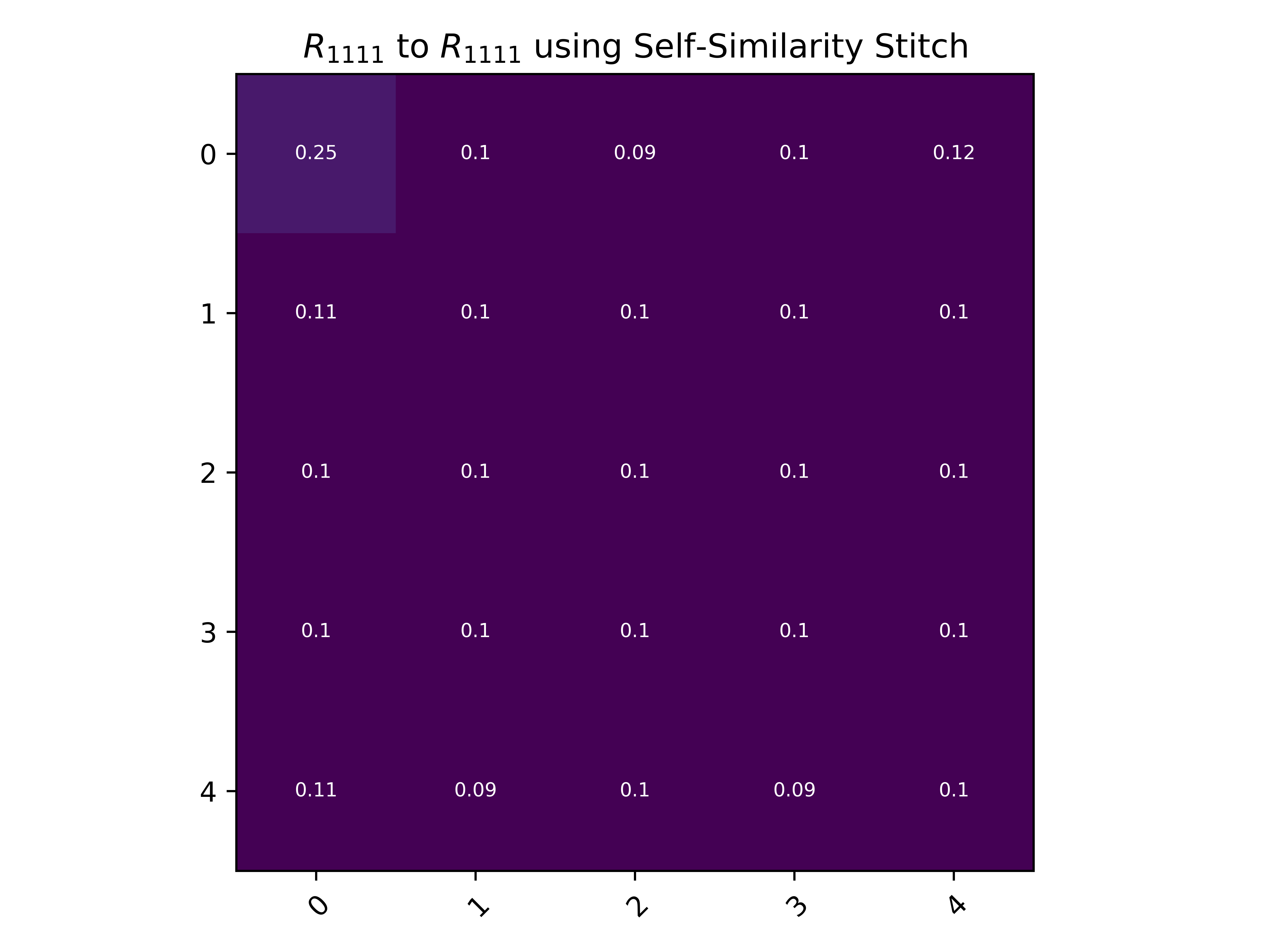}
     \caption{We are unable to yield high task accuracy for stitched networks using
     similarity-trained stitches. This makes sense, since having no task information, it does
     not know what subspaces to prioritize. Most likely, only a few subspaces and/or weights
     truly matter for task accuracy, per the Lottery-Ticket hypothesis \cite{LotteryTicket}. Thus,
     not knowing which those are, the similarity-trained stitch cannot yield high task accuracy
     even when it is trained for thirty epochs to the vanilla stitch's four (the latter yielding
     a very high similarity near 90\% for the lower left-side triangle of the corresponding
     similarity matrix, per the previous figures in this paper).}
  \end{figure}
\end{center}

\subsection*{Image Generation}
\begin{center}
  \begin{figure}[H]
     \centering
     \includegraphics[width=10cm]{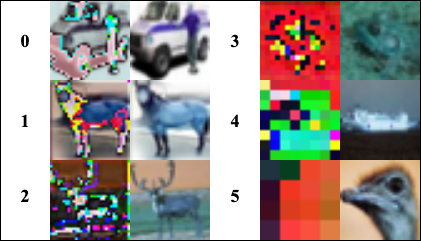}
     \caption{Our results from generating images using stitches.
     We stitched the output of intermediate
     layers (numbered on the left from zero through five to signify the 
     number of blocks coming before it---zero
     corresponding to the output of the first
     convolution, for example) into the very first layer, thereby
     generating images. We hoped to
     understand whether the stitches
     were able to hack the reciever. We did, however,
     not find a discernable pattern other than the loss
     of granularity as the layers progressed.
     On the left side of each pair is the stitch-generated
     image, whereas on the right side is the actual image.}
  \end{figure}
\end{center}
\end{document}